\documentclass[10pt,twocolumn,letterpaper]{article}

\usepackage{iccv}
\usepackage{times}
\usepackage{epsfig}
\usepackage{graphicx}
\usepackage{amsmath}
\usepackage{amssymb}
\usepackage{algorithm}
\usepackage{booktabs}
\usepackage{algorithmic}
\usepackage{dsfont}
\usepackage{multirow}
\usepackage[export]{adjustbox}
\usepackage{textcomp}
\DeclareMathOperator*{\argmax}{arg\,max}

\usepackage{tabularx}

\usepackage{array, boldline, makecell, booktabs}

\usepackage[svgnames, table]{xcolor}

\usepackage[pagebackref=true,breaklinks=true,colorlinks,bookmarks=false]{hyperref}
\iccvfinalcopy 

\begin{document}

\title{DMN4: Few-shot Learning via Discriminative Mutual Nearest Neighbor Neural Network}

\author{Yang Liu \textsuperscript{1}, Tu Zheng\textsuperscript{1}, Jie Song \textsuperscript{2},  Deng Cai\textsuperscript{1}, Xiaofei He\textsuperscript{1} \\
\textsuperscript{1} State Key Lab of CAD\&CG, Zhejiang University,
\textsuperscript{2} Zhejiang Lab \\
{\tt\small \{lyng\_95,sjie\}@zju.edu.cn,} \\ 
{\tt\small \{zhengtuzju,dengcai\}@gmail.com, xiaofeihe@cad.zju.edu.cn}
}

\maketitle
\begin{abstract}
    Few-shot learning (FSL) aims to classify images under low-data regimes, where the conventional pooled global feature is likely to lose useful local characteristics. Recent work has achieved promising performances by using deep descriptors. They generally take all deep descriptors from neural networks into consideration while ignoring that some of them are useless in classification due to their limited receptive field, \eg, task-irrelevant descriptors could be misleading and multiple aggregative descriptors from background clutter could even overwhelm the object's presence. In this paper, we argue that a Mutual Nearest Neighbor (MNN) relation should be established to explicitly select the query descriptors that are most relevant to each task and discard less relevant ones from aggregative clutters in FSL. Specifically, we propose Discriminative Mutual Nearest Neighbor Neural Network (DMN4) for FSL. Extensive experiments demonstrate that our method outperforms the existing state-of-the-arts on both fine-grained and generalized datasets.
\end{abstract}

\section{Introduction}

With the availability of large-scale training data, deep neural networks have achieved great success in recent years \cite{he2016deep,krizhevsky2012imagenet,Simonyan15}. However, collecting and labeling training data are still laboriously painful. In terms of low-data scenarios, such as medical images and endangered species, deep neural networks can easily collapse. Few-shot learning (FSL), whose goal is to construct a model that can be readily adapted to novel classes given just a small number of labeled instances, has emerged as a promising paradigm \cite{sung2018learning,finn2017model,li2019revisiting,munkhdalai2017meta,snell2017prototypical,vinyals2016matching} to alleviate this problem. The main challenge in FSL is to make the best use of accessible labeled data to improve model generalization.
\begin{figure}[t]
    \centering
    \includegraphics[width=0.47\textwidth]{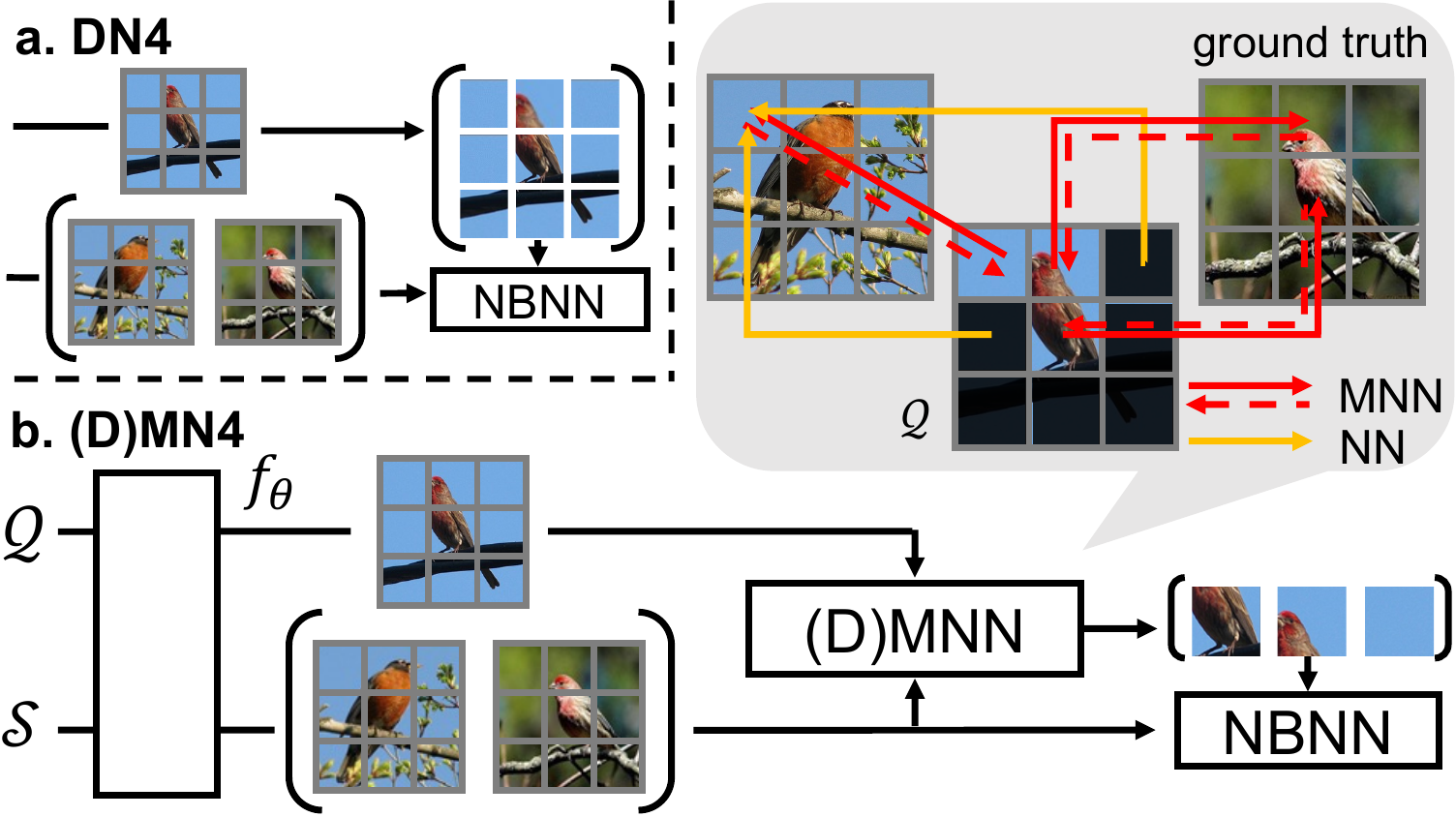}
    \caption{DN4 \cite{li2019revisiting} accumulates all query descriptors $q \in \mathcal{Q}$ where multiple "\textit{sky}" descriptors are taken as strong evidence against \textit{birds}' presence. (D)MN4 selects discriminative task-relevant query descriptors (grids with \textcolor{red}{double-ended red lines}) by introducing (D)MNN where less relevant query descriptors (grids with \textcolor{orange}{orange lines}) nearest neighboring to the same support descriptor would be ignored if they were not the mutual nearest one.}
    \label{fig:intro}
\end{figure}

Few-shot learning methods can be roughly categorized into two schools, \ie, meta-learning based \cite{finn2017model,cai2018memory,munkhdalai2017meta,santoro2016meta} and metric learning based \cite{vinyals2016matching,snell2017prototypical,sung2018learning,li2019revisiting}. Meta-learning aims to accumulate knowledge from learning multiple tasks, and its task-agnostic property makes it easily generalize to new tasks. Metric learning methods mainly focus on concept representation or relation measures by learning a deep embedding space to transfer knowledge. They generally treat deep pooled features from the global average pooling layer as an image-level representation, which is a common practice for large-scale image classification. Considering the unique characteristic of FSL (\ie, the scarcity of examples for each class), however, the cluttered background and large intra-class variations would drive these pooled global representations from the same category far apart in a given metric space under low-data regimes, where useful local characteristics could be overwhelmed and lost.

To fully exploit the local characteristics, Naive-Bayes Nearest Neighbor (NBNN) \cite{Boiman2008In} is recently revisited by DN4 \cite{li2019revisiting} for FSL to retain all reference deep descriptors in their original form. They remove the last global average pooling layer to achieve a dense image representation and treat the output feature map as a set of deep local descriptors. For each descriptor from a query image, they calculate its similarity scores to the nearest neighbor descriptors in each support class. Finally, similarity scores from all query descriptors are accumulated as an image-to-class similarity.

However, in our perspective, there is a notable difference between local invariant descriptors (\eg, SIFT) in traditional NBNN \cite{Boiman2008In} and network deep descriptors \cite{li2019revisiting}: the former one is position-agnostic and diversely distributed (around salient positions), while deep descriptors from neural networks are densely distributed like a grid. Directly accumulating all deep descriptors violates the intuition that the presence of background clutter shouldn't be taken as a strong evidence against the object's presence.

Although the background clutters influence the NBNN classification, there is rarely a straightforward way to pre-select those backgrounds unless introducing extra modules for the foreground retrieval. Instead, we try to mitigate the influences from those descriptors in a different way by recognizing the fact that descriptors within a background clutter are relatively similar to their nearby descriptors, \eg multiple local characterless \textit{blue sky} in Figure \ref{fig:intro} themselves are quite similar compared to the huge difference between the characteristic \textit{beak} and \textit{wings} of the \textit{bird}.

Based on this observation, we introduce Mutual Nearest Neighbor (MNN) for NBNN in this paper to mitigate the accumulated influences from aggregative background clutters so that less relevant characterless background descriptors account less during classification. To further mine the discriminative descriptors in classification, we propose a novel Discriminative Mutual Nearest Neighbor (DMNN) algorithm based on the derivation of NBNN, which is quantitatively shown to be effective in the experiments. In summary, the contributions are: (1) We propose to find discriminative descriptors to improve NBNN based few-shot classification. To the best of our knowledge, this is the first attempt to combine MNN with NBNN in deep learning framework. (2) The proposed methods outperform the state-of-the-art on both fine-grained and generalized few-shot classification datasets. The proposed methods could also be easily extended to a semi-supervised version without extra bells or whistles.

\section{Related Work}

\textbf{Global Representation based methods.} Traditional metric learning based methods use compact feature vectors from the last global average pooling layer of the network to represent images and classification is performed via simple classifiers or nearest neighbors directly. MatchingNet \cite{vinyals2016matching} trains a learnable nearest neighbor classifier with a deep neural network. Prototypical Network \cite{snell2017prototypical} takes the mean of each class as its corresponding prototype representation to learn a metric space. RelationNet \cite{sung2018learning} introduces an auxiliary non-linear metric to compute the similarity score between each query and support set. These deep global representations would lose considerable discriminative local information under low-data regimes.

\textbf{Deep Descriptor based methods.} Another branch of metric learning methods focuses on using deep descriptors to solve few-shot classification. Lifchitz \etal \cite{lifchitz2019dense} proposes to make predictions for each local representation and average their output probabilities. DeepEMD \cite{zhang2020deepemd} adopts the earth mover’s distance as a metric to compute a structural distance between dense image representations to determine image relevance. DN4 \cite{li2019revisiting} uses the top $k$ nearest vectors between two feature maps in a Naive-Bayes way to represent image-level distance. Our (D)MN4 further highlights the importance of selecting discriminative and task-relevant descriptors in the deep descriptors based method.

\textbf{Subspace Learning based methods.} Several recent works also investigate the potential of adaptive subspace learning in FSL. TAPNet \cite{yoon2019tapnet} learns a task-specific subspace projection and the classification is performed based on the mapped query features and projected references. DSN \cite{simon2020adaptive} learns class-specific subspaces based on the few examples within each class, and the classification is performed based on the shortest distance among query projections onto each subspace. Both of them adapt subspace projection with few examples provided in each task but ignore that projection matrices derived from matrix decomposition could easily collapse under low-data regimes. (D)MN4 could also be treated among the family of subspace learning as it also selects a subset of descriptors for each query example. Differently, our subspace dimensionality will be automatically determined by the number of MNN pairs instead of pre-defining a hyper-parameter as in the previous literature \cite{simon2020adaptive, yoon2019tapnet}. Also, the large quantity of descriptors makes it more reliable to retain useful local characteristics compared to the matrix decomposition on a global representation vector.

\section{Methodology}

In this work, we focus on the $N$-way $K$-shot few-shot classification problem, where $N$ is the number of categories with $K$ labeled examples in each. Given a query example, the goal is to classify it into one of the $N$ support classes.

Under this setting, the model is trained with a large training dataset $\mathcal{D}_{train}$ of labeled examples from classes $\mathcal{C}_{train}$ with an episodic training mechanism \cite{vinyals2016matching}. In each episode, we first construct a support set $\mathcal{S}=\{(x_i,y_i)\}_{i=1}^{\textit{N}\times \textit{K}}$ and a query set $\mathcal{Q}=\{(\tilde{x}_i,\tilde{y}_i)\}_{i=1}^q$ containing different samples from the same label, where $\mathcal{S}$ and $\mathcal{Q}$ are sampled from $\mathcal{D}_{train}$; then the model is updated on this small labeled support set $\mathcal{S}$ by minimizing its classification loss on $\mathcal{Q}$. 

\subsection{Deep Descriptor based Image Representation}

We embed an image $x$ via the backbone network to obtain a 3D representation $f_{\theta}(x)\in \mathbb{R}^{C\times H\times W}$, where $f_{\theta}(\cdot)$ is the hypothesis function of the deep backbone network and $\theta$ indicates its parameters. Like other descriptor based methods \cite{krizhevsky2009learning, zhang2020deepemd, lifchitz2019dense}, we treat it as $M$ number of $C$ channel-dimensional descriptors where $M$ is given by $H\times W$.

There are $K$-shot images for each support class within an episode. When $K > 1$, some methods use the empirical mean of $K$ compact image representations for the stability and memory efficient in meta-training. Others instead unite those $K\times r$ feature vectors from the same support class to retain descriptors in their original form. In this work, we use the empirical mean of descriptors that are from the deeper feature extractor (\eg, ResNet-12) while unit in their original form for those that are from the shallower backbone network (\eg, Conv-4).

Formally, we denote the set of descriptors from the same support class $c\in C$ as $\mathbf{s}_c$ and denote descriptors from each query image as $\mathbf{q}$. We use the bold font $\{\mathbf{q},\mathbf{s}\}$ to represent a set of descriptors and $\{q,s\}$ to represent a single channel-dimensional descriptor vector in the following sections. 

\subsection{Mutual Nearest Neighbor}\label{sec:MNN}

As discussed, if we directly accumulate all descriptors in a Naive-Bayes way, background clutters and outliers would mislead the classification. To alleviate it, we revisit the concept of Mutual Nearest Neighbor (MNN) \cite{gowda1979condensed} that initially proposed to obtain a condensed training set decades ago. Formally, we use a single merged support descriptor pool $\mathbf{S} = \bigcup_{c\in C}\mathbf{s}_c$ comprising support descriptors from all classes. For each descriptor $q\in \mathbf{q}$, we find its nearest neighbor $s =\mathrm{NN}_{\mathbf{S}}(q)$ from the support descriptor pool $\mathbf{S}$ and use $s$ to search back its nearest neighbor $\tilde{q} =\mathrm{NN}_{\mathbf{q}}(s)$ from $\mathbf{q}$. If $q$ equals $\tilde{q}$, we consider $q$ and $s$ a MNN pair between query descriptor set $\mathbf{q}$ and support descriptor pool $\mathbf{S}$.

\begin{figure*}[t]
    \centering
    \includegraphics[width=1.0\textwidth]{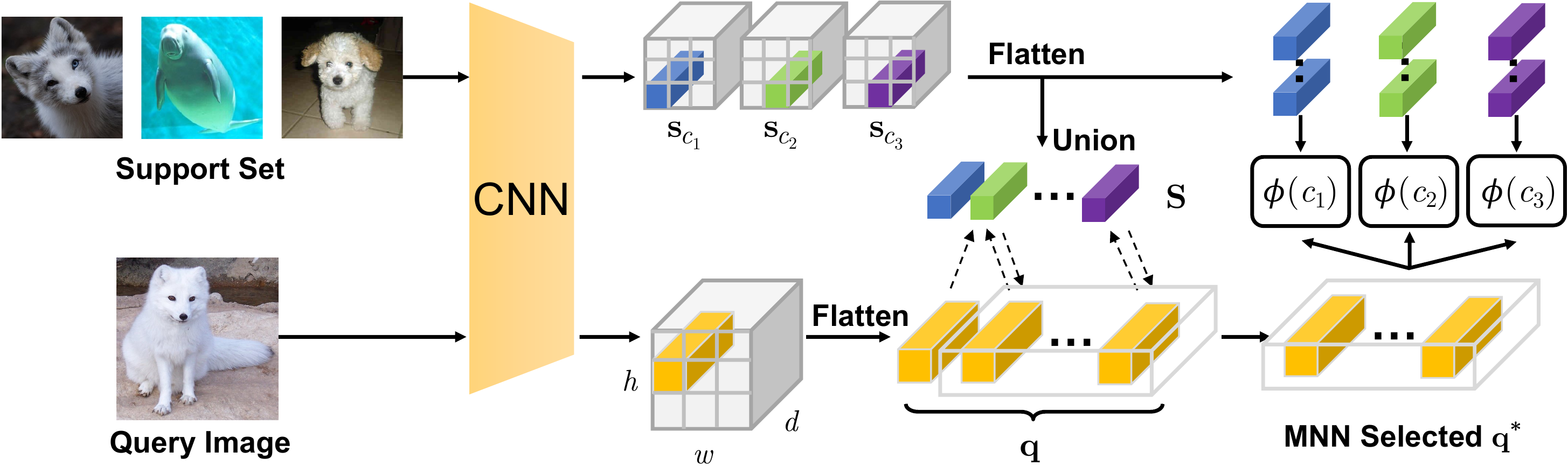}
    \caption{The architecture of \textbf{M}utual \textbf{N}earest \textbf{N}eighbor \textbf{N}eural \textbf{N}etwork (MN4) for few-shot classification. The episodic data is first fed into embedding CNN to get a deep compact feature map and then flatten as sets of channel-dimensional local descriptors $\mathbf{q}$ and $\mathbf{s}_c$. Support descriptors from different classes unite as a single support pool $\mathbf{S}$. We select part of the query descriptors $\mathbf{q}^*$ by performing MNN between $\mathbf{q}$ and $\mathbf{S}$. The selected $\mathbf{q}^*$ as well as the native $\mathbf{s}_c$ are used to calculate a Naive-Bayes classification score $\phi(c)$ for class $c$. (The architecture of DMN4 is the same as MN4 except for using DMNN instead of MNN in selectivity.)}
    \label{fig:arch}
\end{figure*}

The motivation for applying MNN to select relatively task-relevant query descriptors is quite straightforward. Let us consider two descriptors $q$ and $s$ from $\mathbf{q}$ and $\mathbf{S}$ respectively. If $q$ feels that $s$ is its closest descriptor, and $s$ also feels the same, then there exists a feeling of mutual closeness between them, and hence they are likely to represent the same local features. On the other hand, if $s$ feels that $q$ is not such a close descriptor, then even if $q$ feels that $s$ is its closest descriptor, the actual bond of relationship between them is relatively weaker. As yet another possibility, if each feels the other is not close to them, they are not likely to represent the same characteristic. In other words, the strength of such closeness between two descriptors is a function of mutual feelings rather than one-way feeling. By analogy it can be said that the relevance of the local feature $q$ to the provided support features $\mathbf{S}$ can be determined by this mutual nearness. Recall that most characterless backgrounds often aggregate like a clutter, it would be better to filter out less relevant query descriptors in case the accumulation of them dominates the Naive-Bayes classification.

\subsection{Naive-Bayes Nearest Neighbor for FSL}

To help motivate and justify our updates to the original NBNN algorithm, we briefly provide an overview of the original NBNN derivation and its application in DN4 \cite{li2019revisiting}. We start by classifying an image $x$ to class $c$ according to:
\begin{equation}\label{eq:00}
    c=\argmax_{c\in C} p(c|x)
\end{equation}

Applying Bayes' rule with the equal class prior and conditional independence assumptions on Eqn.(\ref{eq:00}) gives:
\begin{align}\label{eq:0}
\begin{split}
\hat{c} &= \argmax_{c \in C} \log(p (x|c)) \\
 &=\argmax_{c \in C} \left[ \log(\prod_{q\in \mathbf{q}} p(q|c)) \right] \\ 
 &=\argmax_{c \in C} \left[ \sum_{q \in \mathbf{q}} \log p(q|c) \right] 
\end{split}
\end{align}

We then approximate $p(q|c)$ in Eqn.(\ref{eq:0}) by a Parzen window estimator with kernel $\kappa$:
\begin{equation}\label{eq:1}
    p(q|c) =\frac{1}{|\mathbf{s}_c|}\sum_{j=1}^{|\mathbf{s}_c|}\kappa(q, \mathrm{NN}_{\mathbf{s}_c}(q, j)) \approx \kappa(q, \mathrm{NN}_{\mathbf{s}_c}(q))
\end{equation}
where $|\mathbf{s}_c|$ is the cardinality of support descriptor set $\mathbf{s}_c$ and $\mathrm{NN}_{\mathbf{s}_c}(q, j)$ is the $j$-th nearest descriptor of support class $c$. NBNN takes it to the extreme by considering only the first nearest neighbor $\mathrm{NN}_{\mathbf{s}_c}(q)$. 

\cite{li2019revisiting} chooses a \textit{cosine similarity} for the approximation of $\log\kappa(\cdot))$ and substitutes Eqn.(\ref{eq:1}) into (\ref{eq:0}) to find the class with the maximum accumulated similarities:
\begin{align}
\begin{split}
    \hat{c} &= \argmax_{c\in C}\left[\sum_{q\in \mathbf{q}}\log\left(\kappa(q, \mathrm{NN}_{\mathbf{s}_c}(q))\right)\right] \\ 
    &\approx \argmax_{c\in C}\left[\sum_{q\in \mathbf{q}}\cos(q, \mathrm{NN}_{\mathbf{s}_c}(q))\right]
\end{split}
\end{align}

In this work, we further selects subspaces $\mathbf{q}^* \in \mathbb{R}^{|\mathbf{q}^*| \times C}$ that owns a relatively stronger bond of mutual closeness with support descriptors. The accumulated similarity score of a query image $x$ to class $c$ in proposed MN4 is 
\begin{equation}
    \phi(x, c)=\sum_{q\in \mathbf{q}^*} \mathrm{cos}(q, \mathrm{NN}_{\mathbf{s}_c}(q))
\end{equation}
and the cross-entropy loss is used to meta-train the network: 
\begin{gather}
       p(c|x) =\frac{e^{\phi(x,c)}}{\sum_{c'\in C}e^{\phi(x,c')}} \\
    \mathcal{L}=-\frac{1}{|\mathcal{Q}|}\sum_{\mathcal{Q}}\sum_{c\in C} y\log p(c|x)
\end{gather}

MNN selects task-relevant descriptors by considering their mutual \textit{absolute} mutual similarity. Yet, it offers no theoretical guarantee that the selected query descriptors are discriminative enough in NBNN classifications. In this section, we propose a novel \textit{relative closeness} in MNN that designed for NBNN classifications and term it Discriminative Mutual Nearest Neighbor (DMNN). The discriminability indicates how query descriptor $q$ relates to its neighbored support descriptor $s$ than other descriptors in $\mathbf{S}$.

We start by recasting NBNN updates as an adjustment to the posterior log-odds \cite{mccann2012local}. Let $c$ be some class and $\bar{c}$ be the set of all other classes, the odds ($\mathcal{O}$) for class $c$ is given by:
\begin{equation}
    \mathcal{O}_c=\frac{p(c|x)}{p(\bar{c}|x)}=\frac{p(x|c)p(c)}{p(x|\bar{c})p(\bar{c})}=\prod_{q\in\mathbf{q}}\frac{p(q|c)}{p(q|\bar{c})}\frac{p(c)}{p(\bar{c})}
\end{equation}

This allows an alternative classification rule expressed in terms of log-odds increments and class priors:
\begin{equation}\label{eq:A}
\hat{c} =\argmax_{c\in C}\left[ \sum_{q\in \mathbf{q}}\log\frac{p(q|c)}{p(q|\bar{c})}+\log\frac{p(c)}{p(\bar{c})}\right]
\end{equation}

Approximating by a Parzen window estimator like in Eqn.(\ref{eq:1}) and assuming an equal class prior give the NBNN log-odds classification rule (find the class with the largest accumulated relative similarities):
\begin{equation} \label{eq:C}
    \hat{c}=\argmax_{c\in C}\sum_{q\in \mathbf{q}} \left(\mathrm{cos}(q, \mathrm{NN}_{\mathbf{s}_c}(q)) - \mathrm{cos}(q, \mathrm{NN}_{\mathbf{S}\setminus \mathbf{s}_c}(q)) \right)
\end{equation}
where $\mathbf{S} \setminus \mathbf{s}_c$ represents all support descriptors set minus the descriptors from class $c$. 

\begin{figure}[t]
    \centering
    \includegraphics[width=0.30\textwidth]{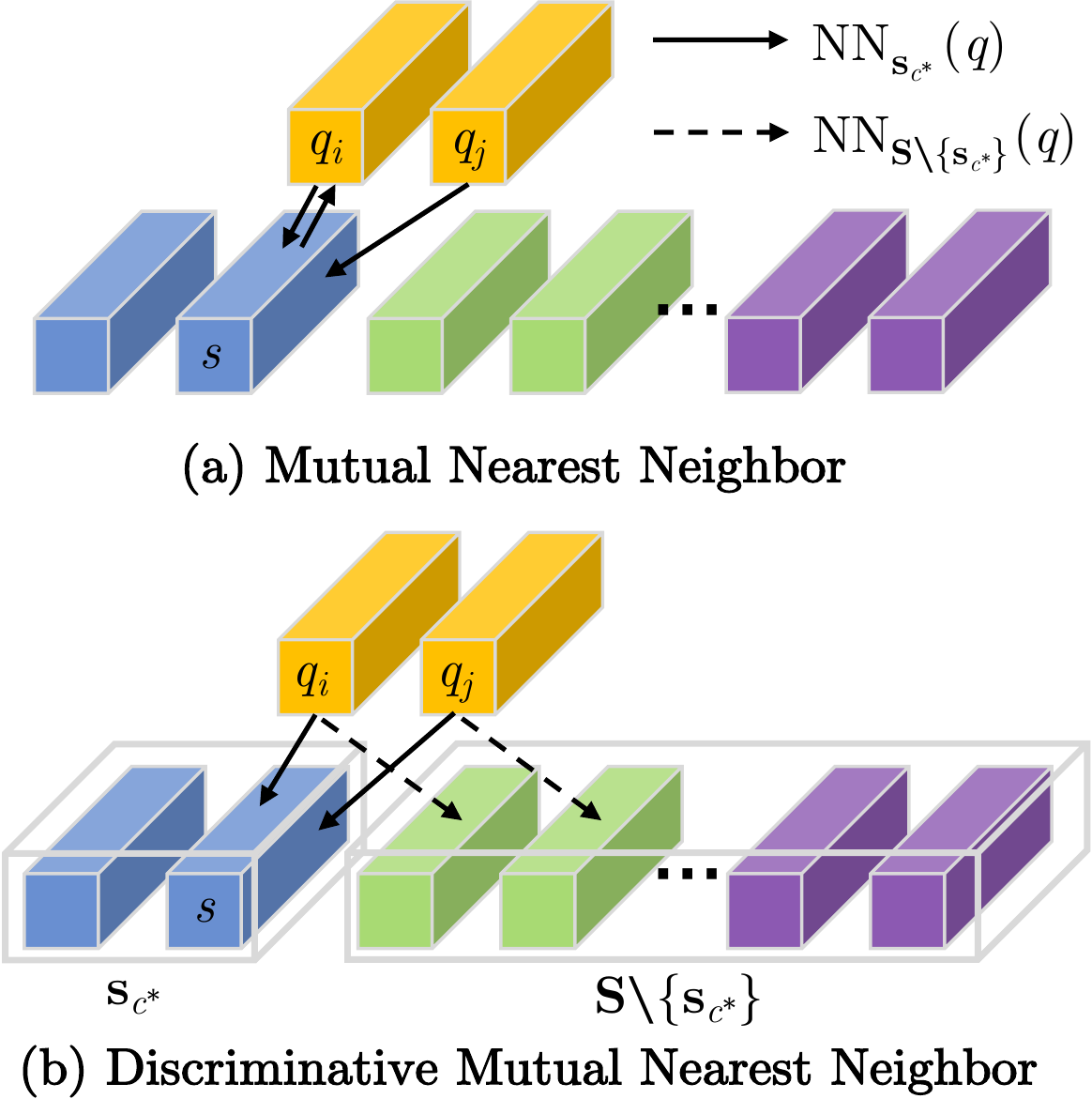}
    \caption{Comparison between MNN and DMNN when multiple query descriptors nearest neighboring to the same support descriptor. (a) MNN selects $q$ by its \textit{absolute} similarity. (b) DMNN selects $q$ by its largest \textit{relative} similarity.}
    \label{fig:DMNNvsMNN}
\end{figure}

Recall that the basic idea of MNN is equivalent to discarding quantities of characterless descriptors. To further guarantee the discriminability of selected descriptors, we take their relative closeness into consideration. Formally, for each query descriptor $q\in \mathbf{q}$, we first find the belonging class $c^*$ of $s\in \mathbf{S}$ that is nearest to $q$, \ie, $\mathbf{s}_{c^*}\owns s=\mathrm{NN}_\mathbf{S}(q)$. To measure whether a query descriptor $q$ discriminative enough in MNN selection, we consider its relative closeness $\tau(q)$ that represents how $q$ votes for its nearest support class $c^*$ than the other supporting classes $C\setminus \{c^*\}$:
\begin{gather}
    c^* = \argmax_{c\in C} \mathrm{cos}(q, \mathrm{NN}_{\mathbf{s}_c}(q)) \label{eq:c_star} \\
    \tau(q) = \mathrm{cos}(q, \mathrm{NN}_{\mathbf{s}_{c^*}}(q)) - \mathrm{cos}(q, \mathrm{NN}_{\mathbf{S}\setminus \mathbf{s}_{c^*}}(q)) \label{eq:tau}
\end{gather}

As illustrated in Figure \ref{fig:DMNNvsMNN}, if both query descriptors $q_i$, $q_j$ are nearest neighboring to the same support descriptor $s$ in support descriptor pool $\mathbf{S}$, the selectivity is determined by their relative closeness $\tau(q)$ in DMNN while determined by the absolute closeness $\mathrm{cos}(q, s)$ in MNN.

\begin{table*}[t]
\centering
\small
\setlength{\tabcolsep}{3.7pt}
\renewcommand\arraystretch{1.2}
\begin{tabular}{ccccccccc}
\hlineB{2.5}
\multicolumn{1}{c}{\multirow{3}{*}{Method}} & \multicolumn{4}{c}{\textbf{Conv-4}}                                             & \multicolumn{4}{c}{\textbf{ResNet-12}} \\
\multicolumn{1}{c}{}                        & \multicolumn{2}{c}{\textit{mini}ImageNet} & \multicolumn{2}{c}{\textit{tiered}ImageNet} & \multicolumn{2}{c}{\textit{mini}ImageNet} & \multicolumn{2}{c}{\textit{tiered}ImageNet} \\ \cmidrule(lr){2-3} \cmidrule(lr){4-5} \cmidrule(lr){6-7} \cmidrule(lr){8-9} 
\multicolumn{1}{c}{}                        & 1-shot          & 5-shot         & 1-shot           & 5-shot          & 1-shot          & 5-shot         & 1-shot           & 5-shot          \\ \hline
MatchingNet \cite{vinyals2016matching} & 43.56 & 55.31 & - & - & 63.08 & 75.99 & 	68.50 & 80.60 \\ 
ProtoNet$^\dagger$ \cite{snell2017prototypical} & 52.32 & 69.74 & 53.19 & 72.28 & 62.67 & 77.88 & 68.48 & 83.46 \\
RelationNet$^\dagger$ \cite{sung2018learning} & 52.12 & 66.90 & 54.33 & 69.95 & 60.97 & 75.12 & 64.71 & 78.41 \\ 
MetaOptNet \cite{lee2019meta} & 52.87 & 68.76 & 54.71 & 71.76 & 62.64 & 78.63 & 65.99 & 81.56 \\ 
DC \cite{lifchitz2019dense} & 49.84 &  69.64 & - & - & 62.53 & 79.77 & - & - \\
TAPNet \cite{yoon2019tapnet} & - & - & - & - &  61.65 & 76.36 &  63.08 & 80.26 \\
DN4$^\dagger$ \cite{li2019revisiting} & 54.66 & 72.92 & 56.86 & 72.16 & 65.35 & 81.10 & 69.60 & 83.41 \\
DSN$^{\nabla\bigstar}$ \cite{simon2020adaptive} & 51.78 & 68.99 & 53.22 & 71.06 & 62.64 & 78.83 & 67.39 & 82.85 \\
DeepEMD$^{\nabla\diamondsuit}$ \cite{zhang2020deepemd} & 52.15 & 65.52 & 50.89 & 66.12 & 65.91 & 82.41 & 71.16 & 86.03 \\
Negative Margin$^\diamondsuit$ \cite{liu2020negative} & 52.84 & 70.41 & - & - & 63.85 & 81.57 & - & - \\
Meta-Baseline \cite{chen2020new} & - & - & - & - & 63.17 & 79.26 & 68.62 & 83.29 \\
Centroid$^\diamondsuit$ \cite{afrasiyabi2020associative} & 53.14 & 71.45 & - & - & 59.88 & 80.35 & 69.29 & 85.97 \\
FEAT \cite{ye2020few} & 55.15 & 71.61 & - & - & 66.78 & 82.05 & 70.80 & 84.79 \\
E$^3$BM$^{\flat\diamondsuit}$ \cite{liu2020ensemble} & 53.20 & 65.10 & 52.10 & 70.20 & 64.09 & 80.29 & 71.34 & 85.82 \\
RFS-Simple \cite{tian2020rethinking} & 55.25 & 71.56 &  56.18 & 72.99 & 62.02 & 79.64 & 69.74 & 84.41 \\
RFS-Distill$^\flat$ \cite{tian2020rethinking} & \textbf{55.88} & 71.65 & 56.76 & 73.21 & 64.82 & 82.14 & 71.52 & \textbf{86.03} \\
FRN$^{\nabla\bigstar}$ \cite{wertheimer2021few} & 54.87 & 71.56 & 55.54 & \textbf{74.68} & 66.45 & 82.83 & \textbf{72.06} & \textbf{86.89} \\
\hline
MN4 (ours) & 55.57 & \textbf{73.64} & \textbf{57.01} & 73.74 & \textbf{66.53} & 83.39 & 71.95 & 85.66 \\
DMN4 (ours) & \textbf{55.77} & \textbf{74.22} & \textbf{56.99} & \textbf{74.13} & \textbf{66.58} & \textbf{83.52} & \textbf{72.10} & 85.72 \\
\hlineB{2.5}
\end{tabular}
\vspace{5pt}
\caption{Few-shot classification accuracy (\%) on \textit{mini}ImageNet and \textit{tiered}ImageNet dataset with Conv-4/ResNet-12 backbones. We show top two performances in \textbf{bold} font regardless of their different settings ($\dagger$: our reimplementation under the same setting. $\nabla$: the reimplemented results with their provided codes on Conv-4. $\diamondsuit$: use SGD fine-tuning during evaluation. $\flat$: knowledge distillation or model ensemble. $\bigstar$: larger shot training). The confidence intervals for our models are all below 0.25. \label{tab:sota}}
\end{table*}

\section{Experiments}

\subsection{Datasets}

\noindent\textbf{\textit{mini}ImageNet} \cite{vinyals2016matching} is a subset of ImageNet containing randomly selected 100 classes. We follow the setup provided by \cite{sachin2017optimization} that takes 64, 16 and 20 classes for training, validation and evaluation respectively.

\noindent\textbf{\textit{tiered}Imagenet} is a larger subset of ImageNet but contains a broader set of classes compared to the \textit{mini}ImageNet. There are 351 classes from 20 different categories for training, 97 classes from 6 different categories for validation, and 160 classes from 8 different categories for testing \cite{ren2018meta}, where the information overlap between training and validation/testing tasks is minimized.

\noindent\textbf{Caltech-UCSD Birds-200-2011 (CUB)} \cite{welinder2010caltech} is a fine-grained dataset that contains 11788 images of 200 birds species. Following the same partition proposed by \cite{hilliard2018few}, we use 100/50/50 classes for training, validation and evaluation respectively. As is commonly implemented, all images are cropped and resized with the provided bounding boxes.

\noindent\textbf{meta-iNat} \cite{wertheimer2019few} is a fine-grained benchmark of animal species in the wild. We follow the class split proposed by where 908 classes of between 50 and 1000 images are used for training and the rest 227 are assigned for evaluation. 

\noindent\textbf{\textit{tiered} meta-iNat} \cite{wertheimer2019few} is a more difficult version of meta-iNat where a large domain gap is introduced between train and test classes. We follow the same class split provided by FRN \cite{wertheimer2021few} where 781/354 classes are used for training and evaluation respectively. 

\subsection{Experimental Settings}

\noindent\textbf{Backbone Networks.} We conduct experiments on both Conv-4 and ResNet-12 backbones. Like in DN4, the Conv-4 generates a feature map of size $19\times19\times 64$ (\ie, 361 deep descriptors of 64 dimensions) for $84\times 84$ image while ResNet-12 gives 25 deep descriptors of 512 dimensions. 

\noindent\textbf{Training and Evaluation.} We meta-train Conv-4 from scratch for 30 epochs by Adam optimizer with learning rate $1\times 10^{-3}$ and decay 0.1 every 10 epochs. With regard to ResNet-12, we first pre-trained it like in the previous literature and then meta-train it by momentum SGD for 40 epochs. The learning rate in meta-training is set $5\times 10^{-4}$ for ResNet-12 and decay 0.5 every 10 epochs.

\noindent\textbf{Evaluation.} We randomly sample 10,000 episodes from the test set during the evaluation and take top-1 mean accuracy criterion and repeat this process 10 times. 

\begin{table*}[t]
\centering
\small
\setlength{\tabcolsep}{8pt}
\renewcommand\arraystretch{1.2}
\begin{tabular}{ccccccc}
\hlineB{2.5}
\multirow{2}{*}{Method} & \multicolumn{2}{c}{CUB} & \multicolumn{2}{c}{meta-iNat} & \multicolumn{2}{c}{\textit{tiered} meta-iNat} \\
                        & 1-shot     & 5-shot     & 1-shot        & 5-shot        & 1-shot            & 5-shot           \\ \hline
ProtoNet$^\heartsuit$ \cite{snell2017prototypical} &      63.73  & 81.50  & 55.34 &  76.43    &  34.34   & 57.13  \\
Covar. pool$^\heartsuit$  \cite{wertheimer2019few} &  -   & -  & 57.15 & 77.20 & 36.06 & 57.48 \\
DSN$^\heartsuit$  \cite{simon2020adaptive}    &   66.01 & 85.41 & 58.08 & 77.38 & 36.82 & 60.11   \\
CTX$^\heartsuit$  \cite{doersch2020crosstransformers} &    69.64  & 87.31  &  60.03 & 78.80 & 36.83 & 60.84 \\
DN4$^\dagger$ \cite{li2019revisiting} & 73.42 & 90.38    &  62.32 & 79.76 & 43.82   &  64.17 \\ 
FRN \cite{wertheimer2021few} &    73.48 & 88.43  &  62.42 & \textbf{80.45} & 43.91 & 63.36 \\
\hline
MN4 (ours)   & \textbf{78.10} & \textbf{92.14}   &    \textbf{62.87} & 80.22 &     \textbf{43.96} &      \textbf{66.93}   \\
DMN4 (ours) &  \textbf{78.36} & \textbf{92.16}      &   \textbf{63.00}  &  \textbf{80.58} &   \textbf{44.10}  & \textbf{67.18} \\ \hlineB{2.5}
\end{tabular}
\vspace{5pt}
\caption{Comparisons of 5-way few-shot classification (\%) results on fine-grained datasets using Conv-4 backbone. $\heartsuit$ indicates results reported by FRN \cite{wertheimer2021few}. The confidence intervals are all below 0.25.\label{tab:finegrained}}
\end{table*}

\begin{table*}[t]
\newcommand{\tabincell}[2]{\renewcommand{\arraystretch}{0.8}\begin{tabular}{@{}#1@{}}#2\end{tabular}}
\centering
\small
\setlength{\tabcolsep}{6pt}
\renewcommand\arraystretch{1.2}
\begin{tabular}{c|ccc|ccc|ccc}
\multirow{2}{*}{} & \multicolumn{3}{c|}{\normalsize{(a) different network depth}} & \multicolumn{3}{c|}{\normalsize{(b) fix informative quality $d$}} & \multicolumn{3}{c}{\normalsize{(c) fix descriptor quality $r$}} \\
                  &   \tabincell{l}{\footnotesize{$r=25$}\\ \footnotesize{$d=640$}} & \tabincell{l}{\footnotesize{$r=100$}\\ \footnotesize{$d=320$}} & \tabincell{l}{\footnotesize{$r=400$}\\ \footnotesize{$d=160$}}  &    \tabincell{l}{\footnotesize{$r=25$}\\ \footnotesize{$d=640$}}
                  & \tabincell{r}{\footnotesize{$r=100$}\\ \footnotesize{$d=640$}}     
                  & \tabincell{r}{\footnotesize{$r=400$}\\ \footnotesize{$d=640$}}  & \tabincell{l}{\footnotesize{$r=100$}\\ \footnotesize{$d=64$}}
                  & \tabincell{r}{\footnotesize{$r=100$}\\ \footnotesize{$d=160$}}     
                  & \tabincell{r}{\footnotesize{$r=100$}\\ \footnotesize{$d=320$}} \\ \hlineB{2.5}
DN4 & 65.35 & 61.73   & 57.00   &  65.35  & 62.60   & 63.00  & 59.16  & 61.15   &   61.73  \\
MN4 & 66.53 & 62.80 & 58.02  & 66.53 & 63.92 & 64.13  & 59.22 & 61.89 & 62.80 \\
DMN4 & 66.58 & 62.94 & 58.73 &  66.58 & 64.32 & 64.77  &  59.25 & 62.23 & 62.94         
\end{tabular}
\caption{Ablations (5-way 1-shot \textit{mini}ImageNet tasks) on different embedding backbones derived from ResNet-12 by: (a) remove entire residual blocks to get larger number $r$ of deep descriptors but less dimensions $d$; (b) remove max pooling layers within some residual blocks but fix feature dimensions $d$; (c) increase feature dimensions $d$ but fix the descriptor's quantity $r$.\label{tab:fm_size}}
\end{table*}

\subsection{Few-shot Classification Results}\label{s:4.3} 

\textbf{\indent Comparisons with the state-of-the-arts.} Table \ref{tab:sota} shows that (D)MN4 achieve new state-of-the-art with simple Conv-4 backbone and have competitive performances when using deeper ResNet-12. (D)MN4 leverages pre-training (in ResNet-12) but no other extra techniques or tricks like inference-time gradient fine-tuning, model ensembling and knowledge distillation.

\textbf{Comparisons with global feature based methods.} Descriptors based methods (\eg, DN4, DC and DeepEMD) generally outperform classic metric-based methods that rely on the image-level feature vector (\eg, MatchingNet, ProtoNet and RelationNet) by a large margin, which validates the effectiveness of using deep descriptors. 

\textbf{Comparisons with descriptor based methods.} Among those methods, DN4 performs NBNN to represent image-level distance; DC averages predictions from each local descriptor; DeepEMD uses optimal matching to connect query and support descriptors. They all use the entire descriptor set while ignoring that some of them are not such discriminative. Our (D)MN4 outperform other model variants on almost all tasks as we think it meaningless to consider a descriptor if it is not task-relevant enough.

\textbf{Comparisons with subspace methods.} Table \ref{tab:sota} shows that class-specific subspace learning (DSN) outperforms task-specific learning (TAPNet). A possible explanation is that, compared to limited class-specific subspaces, there are far more possible variants of task-specific subspaces from different class combinations, where projection matrices could easily collapse under low-data regimes. In contrast, our (D)MN4 outperforms previous methods by (1) using MNN relations to find subspaces where local characteristics are retained in their original forms comparing to matrices decomposition used in DSN; (2) using a set of deep descriptors instead of a single vector representation to avoid model collapsing.

\textbf{Comparisons on fine-grained datasets.} The fine-grained few-shot classification results are shown in Table \ref{tab:finegrained}. It can be observed that our proposed (D)MN4 are superior across the board. Interestingly, our methods achieve overwhelming performances on CUB with the simple Conv-4 backbone. The reason is that cropped and resized images in CUB have few background clutters where MNN relations can be easily established among local characteristics, \eg, \textit{eyes} and \textit{beak}.

\subsection{Ablation Study}

\subsubsection{Different Network Generated Descriptors.} 

It can be observed in Table \ref{tab:sota} that MN4 has greater improvement over DN4 with Conv-4 backbone compared to ResNet-12 and notice that descriptors from ResNet-12 ($r=25,d=640$) are much scarcer but more informative than those from Conv-4 ($r=361,d=64$). We speculate that different quantities and informative quality of descriptors benefit differently from the mutual nearest neighbor selectivity. 

To validate and further investigate the benefit of proposed methods for different kinds of network generated descriptors, we conduct ablations on different embedding backbones derived from ResNet-12. Table \ref{tab:fm_size}(a) firstly shows descriptors from various network depths where deeper embedding networks have less improvement with MNN. It supports the intuition that descriptors from a deep backbone own a large receptive field and contain compact image information where ignoring part of them could be helpless. Table \ref{tab:fm_size}(bc) shows the impact of descriptor quantity $r$ and informative quality $d$ by controlling variables. It can be concluded that MNN have a larger benefit when there are more deep descriptors and more information contained among them. Combining Table \ref{tab:fm_size}(a) with Table \ref{tab:fm_size}(b)(c), we can also conclude that the influence of quantity $r$ is larger than the informative quality $d$ of deep descriptors.

\begin{figure}[t]
    \centering
    \includegraphics[width=0.50\textwidth,right]{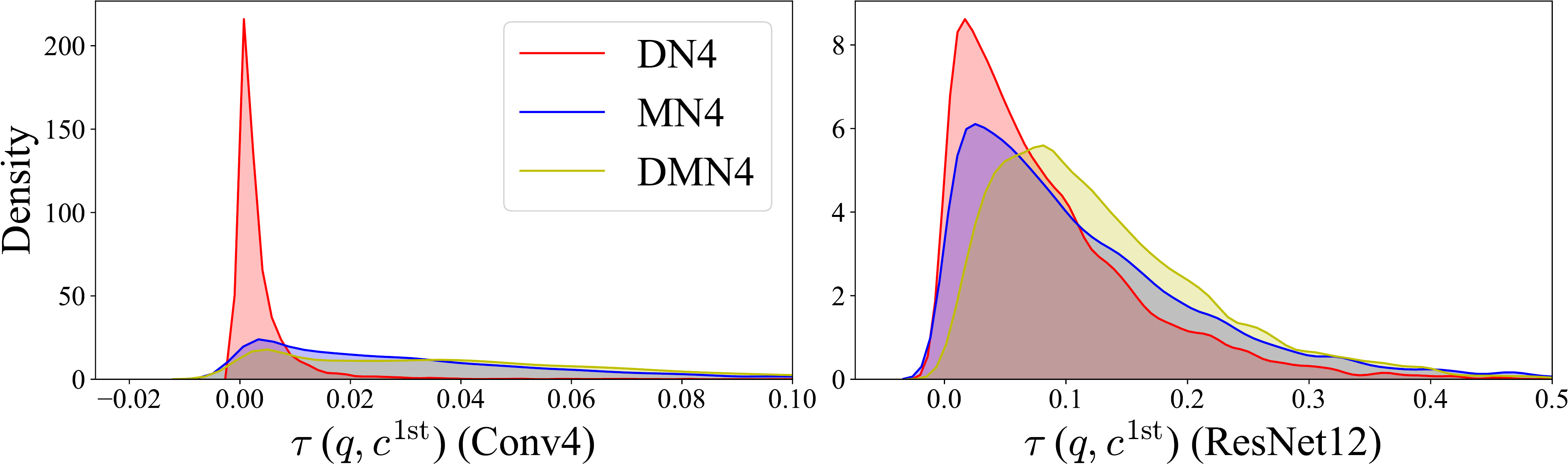}
    \caption{Kernel Density Estimation (KDE) of $\tau(q)$ from sampled query descriptors in three methods.}
    \label{fig:quality_terms}
\end{figure}

\begin{table}[t]
\small
\centering
\setlength{\tabcolsep}{4pt}
\renewcommand\arraystretch{1.2}
\begin{tabular}{cccccc}
\hlineB{2.5}
\multirow{2}{*}{} & \multicolumn{4}{c}{DN4 with ODM} & \multirow{2}{*}{MN4 \footnotesize{(\textit{k}\%)}} \\ \cline{2-5}
                  & 30\%    &  25\%      &  20\%  & 15\% & \\ \hlineB{2.5}
CUB         & 77.02 & 77.31 & 76.19 & 77.23 & \textbf{78.10} \footnotesize{(20.9\%)} \\
\textit{tiered}ImageNet & 56.66 & 56.58 & 56.23 & 55.97 & \textbf{57.01} \footnotesize{(20.0\%)} \\
\textit{mini}ImageNet & 54.40 & 53.76 & 53.99 & 53.73 & \textbf{55.57} \footnotesize{(25.0\%)} \\
\hlineB{2.5}
\end{tabular}
\vspace{5pt}
\caption{Classification accuracy (\%) of DN4 (Conv-4) with Online Discriminative Mining (ODM) on the 5-way 1-shot tasks. We report the averaged percentage (\textit{k}\%) of selected query descriptors in MN4 for comparisons.\label{tab:dn4_ohem}}
\end{table}

\subsubsection{Quality of Selected Descriptors in (D)MN4.} 
\begin{figure}[t]
    \centering
    \begin{minipage}{0.235\textwidth}
    \centering
    \includegraphics[width=1.0\textwidth]{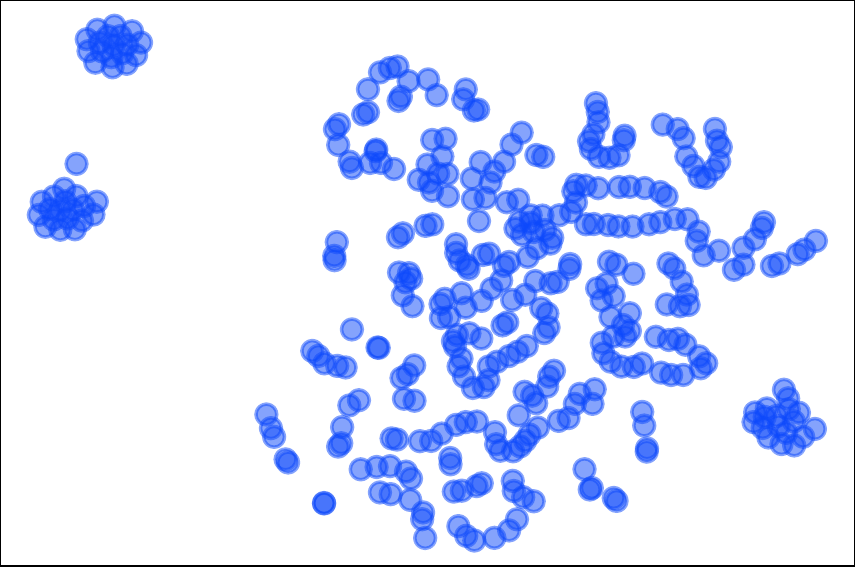}
    \end{minipage}%
    \begin{minipage}{0.235\textwidth}
    \centering
    \includegraphics[width=1.0\textwidth]{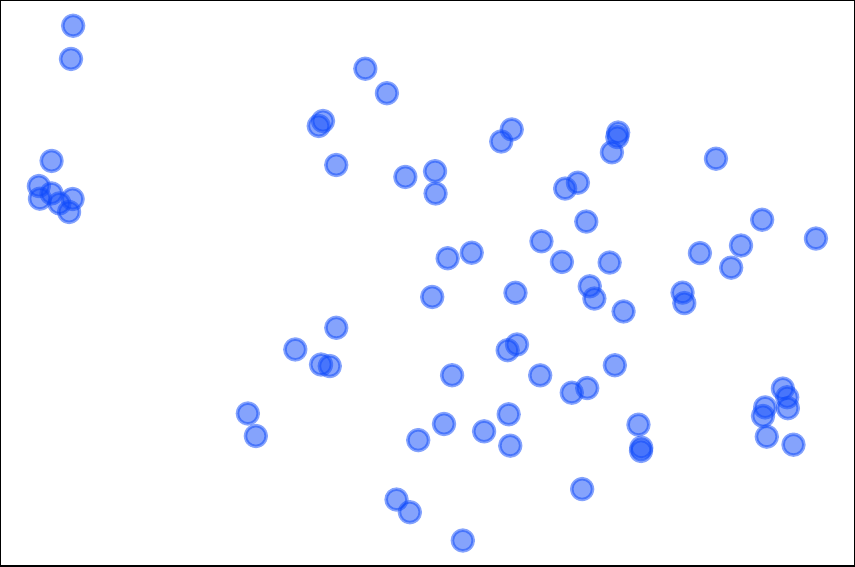}
    \end{minipage} \\
    \vspace{3pt}
    \begin{minipage}{0.235\textwidth}
    \centering
    \footnotesize{$\mathbf{q}$}
    \end{minipage}%
    \begin{minipage}{0.235\textwidth}
    \centering
    \footnotesize{$\mathbf{q}^*$}
    \end{minipage}%
    \caption{The t-SNE visualization of MN4 selected descriptors on an example of \textit{mini}ImageNet with Conv-4 backbone.\label{fig:t-sne}}
\end{figure}

We have claimed that query descriptor $q\in \mathbf{q}^*$ that mutual nearest neighbor to some support descriptors contains class-specific information. To validate, we visualize the $\mathbf{q}$ and selected $\mathbf{q}^*$ with t-SNE in Figure \ref{fig:t-sne}. It can be observed that the visualization of $\mathbf{q}^*$ is departed while $\mathbf{q}$ is a mess.

We also claim (D)MN4 being able to find discriminative query descriptors in this work. To investigate the definition of such discriminability, we further recast NBNN to the log-odds updates of each class and find the relative similarity $\tau(q)$ in Eqn.(\ref{eq:tau}) can be a good measure. To measure the discriminative quality of selected descriptors, we conduct the experiment by randomly sampling 50K query descriptors from DN4, MN4 and DMN4 respectively on \textit{mini}ImageNet and visualize their kernel density estimations (KDE) of $\tau(q)$ in Figure \ref{fig:quality_terms}. It can be found that most descriptors in DN4 contribute little in classification ($\tau(q)\approx 0$) which verifies our claim that not all descriptors are task-relevant. We also find the KDE of Conv-4 backbone is much steeper than that of ResNet-12 indicating deep compact descriptors in ResNet-12 are generally informative and useful. Overall, $\tau(q)$ are much larger in DMN4 revealing that more discriminative descriptors are selected and sampled.

\subsubsection{Why Still Need Mutual Nearest Neighbor?} We have claimed that $\tau(q)$ is a good measure of discriminative effect for the query descriptors and our goal is to find such descriptors. Thus, it is straightforward to raise a solution by selecting the top $k$\% query descriptors of largest $\tau(q)$ like OHEM \cite{shrivastava2016training}. To compare, we conduct an experiment by choosing top $[30\%, 25\%, 20\%, 15\%]$ query descriptors for NBNN classification and the results are shown in Table \ref{tab:dn4_ohem}. It can be observed that \textit{Online Discriminative Mining} (ODM) could definitely improve the classification accuracy, however, MN4 still outperforms them in all tasks. We speculate that this improvement is due to MNN being able to preserve $\textit{rank}$ (\ie, variety) of selected descriptors, where aggregative query descriptors would be ignored if they are nearest neighbors to the same support descriptor but not neighbored back from it. In contrast, ODM only focuses on the top discriminative query descriptors but ignores that similar descriptors (\eg, adjacent background descriptors) would be all retained if they were discriminative enough. To validate, we conduct a rank accuracy experiment by replacing the absolute similarity scores with (rank) counts:
\begin{equation}\label{eq:quantity_only}
    \hat{c}=\argmax_c\sum_{q\in \mathbf{q}^*}\mathds{1}(c = c^*)
\end{equation}
where $\mathds{1}$ is an indicator function that equals 1 if its argument is true and zero otherwise. $c^*$ is the nearest supporting class of query descriptor $q$ as defined in Eqn.(\ref{eq:c_star}).  

\begin{table}[t]
\small
\centering
\setlength{\tabcolsep}{3pt}
\renewcommand\arraystretch{1.2}
\begin{tabular}{cccccc}
\hlineB{2.5}
\multicolumn{2}{c}{\multirow{2}{*}{}} & \multicolumn{2}{c}{\textit{mini}ImageNet} & \multicolumn{2}{c}{CUB} \\ \cline{3-6} 
\multicolumn{2}{c}{} & 1-shot          & 5-shot         & 1-shot     & 5-shot     \\ 
\hlineB{2.5}
\multirow{2}{*}{DN4}       & NBNN     & 54.66 & 72.92 & 73.42 & 90.38 \\
                           & Rank     & 48.84$^{\downarrow \mathrm{5.8}}$ & 58.28$^{\downarrow \mathrm{14.6}}$ & 54.80$^{\downarrow \mathrm{18.6}}$& 77.15$^{\downarrow \mathrm{13.2}}$ \\
\hline
DN4   & NBNN     & 54.50 & 72.60 & 76.34 & 92.12 \\
(ODM) & Rank     & 51.42$^{\downarrow \mathrm{3.1}}$ & 69.00$^{\downarrow \mathrm{3.6}}$ & 65.93$^{\downarrow \mathrm{10.4}}$& 88.69$^{\downarrow \mathrm{3.5}}$ \\
\hline
\multirow{2}{*}{MN4}       & NBNN     & 55.57 & 73.64 & 78.10 & 92.28 \\
                           & Rank     & 52.71$^{\downarrow \mathrm{2.9}}$ & 71.53$^{\downarrow \mathrm{2.1}}$ & \textbf{72.20}$^{\downarrow \textnormal{\textbf{5.9}}}$ & \textbf{89.99}$^{\downarrow \textnormal{\textbf{2.1}}}$\\
\hline
\multirow{2}{*}{DMN4}      & NBNN     & 55.77 & 74.22 & 78.36 & 92.11 \\
                           & Rank     & \textbf{53.97}$^{\downarrow \textnormal{\textbf{2.3}}}$ & \textbf{72.60}$^{\downarrow \textnormal{\textbf{1.6}}}$ & 70.58$^{\downarrow \mathrm{7.8}}$& 88.80$^{\downarrow \mathrm{3.3}}$ \\
\hlineB{2.5}  
\end{tabular}
\vspace{5pt}
\caption{Comparisons of NBNN accuracy and rank accuracy (\%) from three models with Conv-4 backbone. For each task, the smallest gap between NBNN and rank accuracy is marked in \textbf{bold}.\label{tab:rank_accuracy}}
\end{table}

It can be observed in Table \ref{tab:rank_accuracy} that the performances of DN4 drop by a large margin if we only count the number of nearest neighbored descriptors in each support class. ODM narrows down the gap by focusing on the top discriminative descriptors. Our (D)MN4 further cuts down the differences by preserving more variety of visual characteristics with mutual nearest relations. More qualitative results in supplementary materials also demonstrate that vast majority of selected query descriptors nearest neighbor to the ground truth class.

\subsection{Semi-Supervised Few-Shot Learning}

From the perspective of MNN that descriptors from an unlabeled image can be roughly categorized to its MNN support class (if exists), our work can be easily extended to \textit{semi}-supervised version MN4-semi as follows: (1) we first pseudo-label each descriptor $u$ from unlabeled images to its MNN support class $c$ and attach it to the support descriptors $\mathbf{s}_c=\{s_{1}^{c}, ...,s_{K\times M}^{c}\}\cup \{u\} $ if their MNN relationship exists. (2) we run MN4 between $\mathbf{S}$ and $\mathbf{q}$ as before. 

We follow the same experimental setup proposed by \cite{ren2018meta} and report the comparisons in Table \ref{tab:semi}, where MN4-semi shows a consistent $\sim 2$\% improvement over the baseline DN4. Also, MN4-semi has a less performance drop compared to DSN \cite{simon2020adaptive} when unlabeled \textit{distractor} classes included as MNN discards outliers from these classes.

\begin{table}[]
\small
\centering
\setlength{\tabcolsep}{5pt}
\renewcommand\arraystretch{1.2}
\begin{tabular}{cccc}
\hlineB{2.5}
\multirow{2}{*}{}      & \multirow{2}{*}{Model} & \multicolumn{2}{c}{5-way Accuracy (\%)} \\ \cline{3-4} 
                       &                        & 1-shot             & 5-shot             \\ \hline
\multirow{6}{*}{w/o D} & PN, Non-Masked \cite{ren2018meta} & 50.09 & 64.59 \\
 & PN, Masked \cite{ren2018meta} & 50.41 & 64.39 \\
 & TPN-semi \cite{liu2019learning} & 52.78 & 66.42 \\
 & DSN-semi \cite{simon2020adaptive} & 53.01& 69.12 \\
 & \textbf{DN4$^\dagger$} \cite{li2019revisiting} & \textbf{51.46} & \textbf{68.75} \\
 & \textbf{MN4-semi (ours)} & \textbf{53.48} & \textbf{71.06} \\ \hlineB{2.5}
\multirow{4}{*}{w/ D}  & PN, Non-masked \cite{ren2018meta} & 48.70 & 63.55 \\
 & PN, Masked \cite{ren2018meta} & 49.04 & 62.96 \\
 & DSN-semi \cite{simon2020adaptive} & 51.01 & 67.12 \\
 & \textbf{MN4-semi (ours)} & \textbf{52.73} & \textbf{70.31}\\ \hlineB{2.5}
\end{tabular}
\vspace{5pt}
\caption{Semi-supervised few-shot classification results using Conv-4 on \textit{mini}ImageNet with 40\% labeled data. We show the classification results (w/ D) and without \textit{distractors} (w/o D). \label{tab:semi}}
\end{table}

\begin{table}[t]
\small
\setlength{\tabcolsep}{3pt}
\centering
\renewcommand\arraystretch{1.2}
\begin{tabular}{ccccc}
\hlineB{2.5}
Method  & Backbone               & Spatial Size      & 1-shot (s) & 5-shot (s) \\ \hline
DeepEMD \cite{zhang2020deepemd} & Conv4$^\dagger \cite{snell2017prototypical}$ & 5$\times$5& 0.173 & 13.05 \\ \hline
DeepEMD \cite{zhang2020deepemd} &   \multirow{4}{*}{Conv4 \cite{li2019revisiting}} & \multirow{4}{*}{19$\times$19} & 70.78 & $\sim$2500 \\
DN4 \cite{li2019revisiting} &   &   & 0.011 & 0.032   \\
\textbf{MN4 (ours)}     & & & 0.015 & 0.053        \\
\textbf{DMN4 (ours)}    & & & 0.012 & 0.052     \\ \hline
DeepEMD \cite{zhang2020deepemd}        & \multirow{4}{*}{ResNet12}      & \multirow{4}{*}{5$\times$5} & 0.235 & 13.26\\
DN4 \cite{li2019revisiting} & & &     0.060 & 0.074 \\
\textbf{MN4 (ours)}  & & &  0.067 & 0.077    \\
\textbf{DMN4 (ours)} & & &    0.080 & 0.077   \\ 
\hlineB{2.5}
\end{tabular}
\vspace{5pt}
\caption{Comparisons of computation time for DN4 \cite{li2019revisiting}, proposed (D)MN4 and DeepEMD \cite{zhang2020deepemd} under 5-way tasks with 15 images per query class. The computation is averaged over 10$^4$ runs.\label{tab:time}}
\end{table}

\section{Time Complexity}

We compare the time complexity with our baseline method DN4 \cite{li2019revisiting} as well as another the state-of-the-art method descriptor based method DeepEMD \cite{zhang2020deepemd} in Table S\ref{tab:time}. The computation time is averaged by 10$^4$ forward inferences. For DeepEMD, we use its open-sourced code deployed with the OpenCV library which achieves the fastest computation speed in their implementation. For DN4, we use our reimplementation for fair comparisons, which is much faster than the original implementation \cite{li2019revisiting}.

It can be observed that (D)MN4 add comparably little extra computation time compared to DN4 but outperforms it on all FSL tasks. DeepEMD has a much larger computation overhead over other Naive Bayes Nearest Neighbor based methods since it solve a linear programming (LP) problem for each forward process. The LP problem in DeepEMD becomes much time-comsuming when the number of deep descriptors grows and it would take hours to run a single forward on 5-shot task with a large spatial size. In contrast, our proposed (D)MN4 is a more portable method of better performances.

\section{Conclusions}

In this paper, we argue that not all deep descriptors are useful in recent few-shot learning methods since task-irrelevant outlier could be misleading and background descriptors could even overwhelm the object's presence. We propose Discriminative Mutual Nearest Neighbor Neural Network (DMN4) to find those that are most task-relevant to each task. Experimental results demonstrate that our method outperforms the previous state-of-the-arts on both supervised and semi-supervised FSL tasks.

\definecolor{cGreen}{RGB}{0,128,0}
\begin{figure*}[t]
    \centering
    \includegraphics[width=1.0\textwidth]{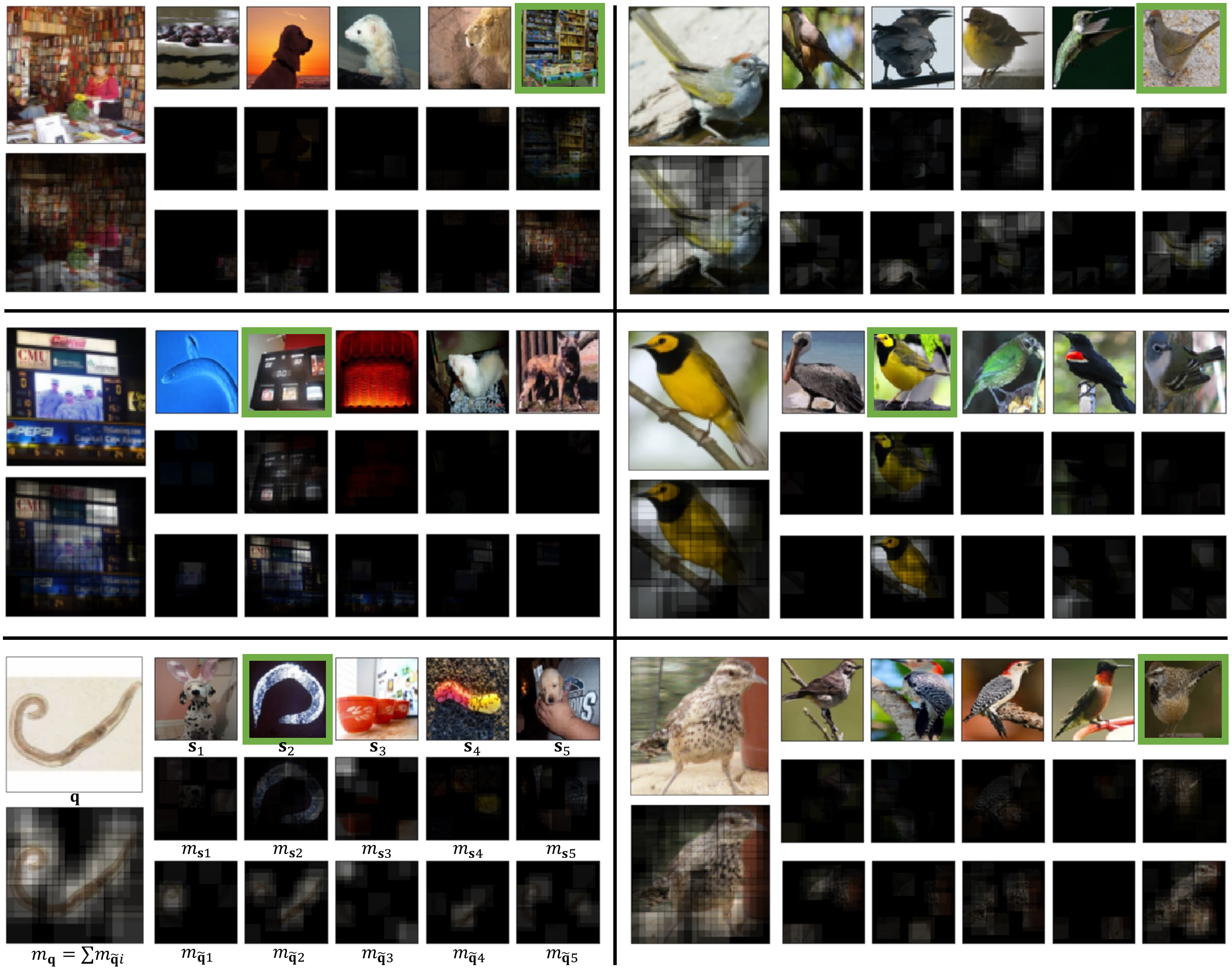}
    \newcolumntype{Y}{>{\centering\arraybackslash}X}
    \begin{tabularx}{\textwidth}{@{}YY@{}}
    \textbf{\textit{mini}ImageNet} & \textbf{Caltech-UCSD Birds-200-2011}
    \end{tabularx}
    \caption{Supplementary visualization of receptive field from different tasks by Conv-4. The ground truth is surrounded by \textcolor{cGreen}{green rectangle}. Examples in left pane are from \textit{mini}ImageNet and those in right pane's are from the fine-grained Caltech-UCSD Birds-200-2011.}
    \label{fig:supp_examples}
\end{figure*}

\section*{Visualization of Selected Descriptors.}

We provide receptive visualization results from different tasks in Figure \ref{fig:supp_examples} to verify that (D)MN4 finds task-relevant descriptors in classification. The qualitative results also demonstrate that vast majority of selected query descriptors nearest neighbor to the ground truth class. We detail our visualization process as follows:

\begin{enumerate}
\item Get descriptor representations $\mathbf{q}$, $\{\mathbf{s}_1, ..., \mathbf{s}_N\}$ for query and support images respectively.
\item Select task-relevant query descriptors $\mathbf{q}^*$ by constructing MNN relations between $\mathbf{q}$ and $\mathbf{S}=\bigcup_{c\in[1,N]}\mathbf{s}_c$. Initialize nearest neighbored descriptor set $\mathbf{s}^*_c=\{\}$ for each class $c$.
\item For each $q\in\mathbf{q}^*$, select $s=\mathrm{NN}_\mathbf{S}(q)$ and record it by $\mathbf{s}^*_c=\mathbf{s}^*_c\cup\{s\}$ where $c$ is the belonging class of $s$.
\item For each descriptors in $\mathbf{q}^*$ and $\{\mathbf{s}^*_1, ..., \mathbf{s}^*_N\}$, the visualization of receptive field is performed \textbf{by the forward and backward passes and looking for the locations with non-zero gradients} in corresponding images. We accumulate the receptive field from different $q\in \mathbf{q}^*$ and normalize it to get $m_\mathbf{q}$ in Figure S\ref{fig:supp_examples}. Similar operations are performed to get $\{m_{\mathbf{s}_1},...,m_{\mathbf{s}_N}\}$ for $\{\mathbf{s}^*_1, ..., \mathbf{s}^*_N\}$ respectively.
\item For each $s\in \mathbf{s}^*_c$, select $\tilde{q}=\mathrm{NN}_\mathbf{q}(s)$ and build mutual nearest query descriptor set $\{\tilde{\mathbf{q}}_1, ..., \tilde{\mathbf{q}}_N\}$ by collecting $\tilde{q}$ of the same supporting class $c$. It can be verified that $\mathbf{q}^*=\bigcup_{c\in[1,N]}\tilde{\mathbf{q}}_c$. We visualize $\{m_{\tilde{\mathbf{q}}_1},...,m_{\tilde{\mathbf{q}}_N}\}$ for $\{\tilde{\mathbf{q}}_1, ..., \tilde{\mathbf{q}}_N\}$ with the same operations as in Step 4.
\end{enumerate}

\clearpage

{\small
\bibliographystyle{ieee_fullname}
\bibliography{egbib}

\begin{thebibliography}{10}\itemsep=-1pt

\bibitem{afrasiyabi2020associative}
Arman Afrasiyabi, Jean-Fran{\c{c}}ois Lalonde, and Christian Gagn{'e}.
\newblock Associative alignment for few-shot image classification.
\newblock In {\em European Conference on Computer Vision}, pages 18--35.
  Springer, 2020.

\bibitem{Boiman2008In}
Oren Boiman, Eli Shechtman, and Michal Irani.
\newblock In defense of nearest-neighbor based image classification.
\newblock In {\em CVPR}, 2008.

\bibitem{cai2018memory}
Qi Cai, Yingwei Pan, Ting Yao, Chenggang Yan, and Tao Mei.
\newblock Memory matching networks for one-shot image recognition.
\newblock In {\em CVPR}, pages 4080--4088, 2018.

\bibitem{chen2020new}
Yinbo Chen, Xiaolong Wang, Zhuang Liu, Huijuan Xu, and Trevor Darrell.
\newblock A new meta-baseline for few-shot learning.
\newblock {\em arXiv preprint arXiv:2003.04390}, 2020.

\bibitem{doersch2020crosstransformers}
Carl Doersch, Ankush Gupta, and Andrew Zisserman.
\newblock Crosstransformers: spatially-aware few-shot transfer.
\newblock {\em arXiv preprint arXiv:2007.11498}, 2020.

\bibitem{finn2017model}
Chelsea Finn, Pieter Abbeel, and Sergey Levine.
\newblock Model-agnostic meta-learning for fast adaptation of deep networks.
\newblock In {\em ICML}, pages 1126--1135. JMLR. org, 2017.

\bibitem{gowda1979condensed}
K Gowda and G Krishna.
\newblock The condensed nearest neighbor rule using the concept of mutual
  nearest neighborhood (corresp.).
\newblock {\em IEEE Transactions on Information Theory}, 25(4):488--490, 1979.

\bibitem{he2016deep}
Kaiming He, Xiangyu Zhang, Shaoqing Ren, and Jian Sun.
\newblock Deep residual learning for image recognition.
\newblock In {\em CVPR}, pages 770--778, 2016.

\bibitem{hilliard2018few}
Nathan {Hilliard}, Lawrence {Phillips}, Scott {Howland}, Artëm {Yankov},
  Courtney~D. {Corley}, and Nathan~O. {Hodas}.
\newblock Few-shot learning with metric-agnostic conditional embeddings.
\newblock {\em arXiv preprint arXiv:1802.04376}, 2018.

\bibitem{krizhevsky2009learning}
Alex Krizhevsky, Geoffrey Hinton, et~al.
\newblock Learning multiple layers of features from tiny images.
\newblock 2009.

\bibitem{krizhevsky2012imagenet}
Alex Krizhevsky, Ilya Sutskever, and Geoffrey~E Hinton.
\newblock Imagenet classification with deep convolutional neural networks.
\newblock In {\em NIPS}, pages 1097--1105, 2012.

\bibitem{lee2019meta}
Kwonjoon Lee, Subhransu Maji, Avinash Ravichandran, and Stefano Soatto.
\newblock Meta-learning with differentiable convex optimization.
\newblock In {\em Proceedings of the IEEE/CVF Conference on Computer Vision and
  Pattern Recognition}, pages 10657--10665, 2019.

\bibitem{li2019revisiting}
Wenbin Li, Lei Wang, Jinglin Xu, Jing Huo, Yang Gao, and Jiebo Luo.
\newblock Revisiting local descriptor based image-to-class measure for few-shot
  learning.
\newblock In {\em CVPR}, pages 7260--7268, 2019.

\bibitem{lifchitz2019dense}
Yann Lifchitz, Yannis Avrithis, Sylvaine Picard, and Andrei Bursuc.
\newblock Dense classification and implanting for few-shot learning.
\newblock In {\em CVPR}, pages 9258--9267, 2019.

\bibitem{liu2020negative}
Bin Liu, Yue Cao, Yutong Lin, Qi Li, Zheng Zhang, Mingsheng Long, and Han Hu.
\newblock Negative margin matters: Understanding margin in few-shot
  classification.
\newblock In {\em European Conference on Computer Vision}, pages 438--455.
  Springer, 2020.

\bibitem{liu2019learning}
Yanbin {Liu}, Juho {Lee}, Minseop {Park}, Saehoon {Kim}, Eunho {Yang}, Sung~Ju
  {Hwang}, and Yi {Yang}.
\newblock Learning to propagate labels: Transductive propagation network for
  few-shot learning.
\newblock In {\em ICLR}, 2019.

\bibitem{liu2020ensemble}
Yaoyao Liu, Bernt Schiele, and Qianru Sun.
\newblock An ensemble of epoch-wise empirical bayes for few-shot learning.
\newblock In {\em European Conference on Computer Vision}, pages 404--421.
  Springer, 2020.

\bibitem{mccann2012local}
Sancho McCann and David~G Lowe.
\newblock Local naive bayes nearest neighbor for image classification.
\newblock In {\em CVPR}, pages 3650--3656. IEEE, 2012.

\bibitem{munkhdalai2017meta}
Tsendsuren Munkhdalai and Hong Yu.
\newblock Meta networks.
\newblock In {\em ICML}, pages 2554--2563. JMLR. org, 2017.

\bibitem{ren2018meta}
Mengye {Ren}, Sachin {Ravi}, Eleni {Triantafillou}, Jake {Snell}, Kevin
  {Swersky}, Josh~B. {Tenenbaum}, Hugo {Larochelle}, and Richard~S. {Zemel}.
\newblock Meta-learning for semi-supervised few-shot classification.
\newblock In {\em ICLR}, 2018.

\bibitem{sachin2017optimization}
Ravi Sachin and Larochell Hugo.
\newblock Optimization as a model for few-shot learning.
\newblock {\em ICLR}, 2017.

\bibitem{santoro2016meta}
Adam Santoro, Sergey Bartunov, Matthew Botvinick, Daan Wierstra, and Timothy
  Lillicrap.
\newblock Meta-learning with memory-augmented neural networks.
\newblock In {\em ICML}, pages 1842--1850, 2016.

\bibitem{shrivastava2016training}
Abhinav Shrivastava, Abhinav Gupta, and Ross Girshick.
\newblock Training region-based object detectors with online hard example
  mining.
\newblock In {\em CVPR}, pages 761--769, 2016.

\bibitem{simon2020adaptive}
Christian Simon, Piotr Koniusz, Richard Nock, and Mehrtash Harandi.
\newblock Adaptive subspaces for few-shot learning.
\newblock In {\em CVPR}, pages 4136--4145, 2020.

\bibitem{Simonyan15}
Karen Simonyan and Andrew Zisserman.
\newblock Very deep convolutional networks for large-scale image recognition.
\newblock In {\em ICLR}, 2015.

\bibitem{snell2017prototypical}
Jake Snell, Kevin Swersky, and Richard Zemel.
\newblock Prototypical networks for few-shot learning.
\newblock In {\em NIPS}, pages 4077--4087, 2017.

\bibitem{sung2018learning}
Flood Sung, Yongxin Yang, Li Zhang, Tao Xiang, Philip~HS Torr, and Timothy~M
  Hospedales.
\newblock Learning to compare: Relation network for few-shot learning.
\newblock In {\em CVPR}, pages 1199--1208, 2018.

\bibitem{tian2020rethinking}
Yonglong Tian, Yue Wang, Dilip Krishnan, Joshua~B Tenenbaum, and Phillip Isola.
\newblock Rethinking few-shot image classification: a good embedding is all you
  need?
\newblock In {\em Computer Vision--ECCV 2020: 16th European Conference,
  Glasgow, UK, August 23--28, 2020, Proceedings, Part XIV 16}, pages 266--282.
  Springer, 2020.

\bibitem{vinyals2016matching}
Oriol Vinyals, Charles Blundell, Timothy Lillicrap, Daan Wierstra, et~al.
\newblock Matching networks for one shot learning.
\newblock In {\em NIPS}, pages 3630--3638, 2016.

\bibitem{welinder2010caltech}
Peter {Welinder}, Steve {Branson}, Takeshi {Mita}, Catherine {Wah}, Florian
  {Schroff}, Serge {Belongie}, and Pietro {Perona}.
\newblock Caltech-ucsd birds 200.
\newblock 2010.

\bibitem{wertheimer2019few}
Davis Wertheimer and Bharath Hariharan.
\newblock Few-shot learning with localization in realistic settings.
\newblock In {\em Proceedings of the IEEE/CVF Conference on Computer Vision and
  Pattern Recognition}, pages 6558--6567, 2019.

\bibitem{wertheimer2021few}
Davis Wertheimer, Luming Tang, and Bharath Hariharan.
\newblock Few-shot classification with feature map reconstruction networks.
\newblock In {\em Proceedings of the IEEE/CVF Conference on Computer Vision and
  Pattern Recognition}, pages 8012--8021, 2021.

\bibitem{ye2020few}
Han-Jia Ye, Hexiang Hu, De-Chuan Zhan, and Fei Sha.
\newblock Few-shot learning via embedding adaptation with set-to-set functions.
\newblock In {\em CVPR}, pages 8808--8817, 2020.

\bibitem{yoon2019tapnet}
Sung~Whan {Yoon}, Jun {Seo}, and Jaekyun {Moon}.
\newblock Tapnet: Neural network augmented with task-adaptive projection for
  few-shot learning.
\newblock In {\em ICML}, pages 7115--7123, 2019.

\bibitem{zhang2020deepemd}
Chi Zhang, Yujun Cai, Guosheng Lin, and Chunhua Shen.
\newblock Deepemd: Few-shot image classification with differentiable earth
  mover's distance and structured classifiers.
\newblock In {\em CVPR}, pages 12203--12213, 2020.

\end{thebibliography}
}

\end{document}